\documentclass[10pt,twocolumn,letterpaper]{article}

\usepackage{cvpr}              

\usepackage{graphicx}
\usepackage{amsmath}
\usepackage{amssymb}
\usepackage{booktabs}
\usepackage{algorithm}
\usepackage{algorithmicx}
\usepackage{algpseudocode}
\usepackage{multirow}
\usepackage{booktabs}
\usepackage{diagbox}
\usepackage[accsupp]{axessibility}  

\usepackage[pagebackref,breaklinks,colorlinks]{hyperref}

\usepackage[capitalize]{cleveref}
\crefname{section}{Sec.}{Secs.}
\Crefname{section}{Section}{Sections}
\Crefname{table}{Table}{Tables}
\crefname{table}{Tab.}{Tabs.}

\def\cvprPaperID{7565} 
\def\confName{CVPR}
\def\confYear{2023}

\begin{document}

\title{Switchable Representation Learning Framework with Self-compatibility}

\author{%
     Shengsen Wu{$^{1,2}$},
    Yan Bai{$^{2}$},
    Yihang Lou{$^{3}$},
    Xiongkun Linghu{$^{4}$}, 
    Jianzhong He{$^{3}$},
    Ling-Yu Duan{$^{1,2}$}
     \and
     {$^{1}$} Peng Cheng Laboratory, China; 
  {$^{2}$} Peking University, China;  \\
  {$^{3}$} Huawei Technologies; 
 {$^{4}$} Tsinghua University, China   \and
     \tt\small \{sswu, yanbai, lingyu\}@pku.edu.cn,\\ 
    \tt\small  \{louyihang1,jianzhong.he\}@huawei.com, 
      lhxk20@mails.tsinghua.edu.cn  
}
\maketitle

\begin{abstract}

Real-world visual search systems involve deployments on multiple platforms with different computing and storage resources.  
Deploying a unified model that suits the minimal-constrain platforms leads to limited accuracy. It is expected to deploy models with different capacities adapting to the resource constraints, which requires features extracted by these models to be aligned in the metric space.
The method to achieve feature alignments is called ``compatible learning''. Existing research mainly focuses on the one-to-one compatible paradigm, which is limited in learning compatibility among multiple models. We propose a \textbf{S}witchable representation learning \textbf{F}ramework with \textbf{S}elf-\textbf{C}ompatibility (SFSC). 
SFSC generates a series of compatible sub-models with different capacities through one training process. 
The optimization of sub-models faces gradients conflict, and we mitigate this problem from the perspective of the magnitude and direction.
We adjust the priorities of sub-models dynamically through uncertainty estimation to co-optimize sub-models properly. Besides, the gradients with conflicting directions are projected to avoid mutual interference.
SFSC achieves state-of-the-art performance on the evaluated datasets.

\end{abstract}

\section{Introduction}

Visual search systems are widely deployed, which recall the nearest neighbors in gallery features according to their distances to the query feature. 
Real-world visual search systems consist of multiple models deployed on different platforms \cite{lou2019front,sandler2018mobilenetv2,graham2021levit} (\emph{e.g.} clouds, mobiles, smart cameras), where different platforms interact with each other by visual features. Typically, as for person re-identification systems, images are captured and processed into features on edge sides. And such features are sent to the cloud side to compare with the database features for 
identifying a specific person by feature similarities.
As diverse platforms meet different computing and storage resource limitations, deploying the unified model which suits the minimal-constrain platforms leads to a waste of resources on other platforms and limited accuracy. To make better use of resources and achieve higher accuracy, it is expected to deploy models with different capacities which adapt to the resource limitations. 
Such a solution requires compatibility among the models to satisfy their interaction requirements, which means that the similar data processed by different models are close to each other in the feature space while dissimilar data are far apart.

To achieve compatibility, compatible learning methods have been proposed. 
Existing research focuses on the one-to-one compatible paradigm \cite{shen2020towards,wu2021neighborhood,budnik2021asymmetric}, which constrains the learned features of a latter (learnable) model to be compatible with its previous (fixed) version. They are limited in achieving many-to-many compatibilities, which means any two in a series of models are compatible with each other. 

\begin{figure*}[htb]
  \centering
   \includegraphics[height=4cm]{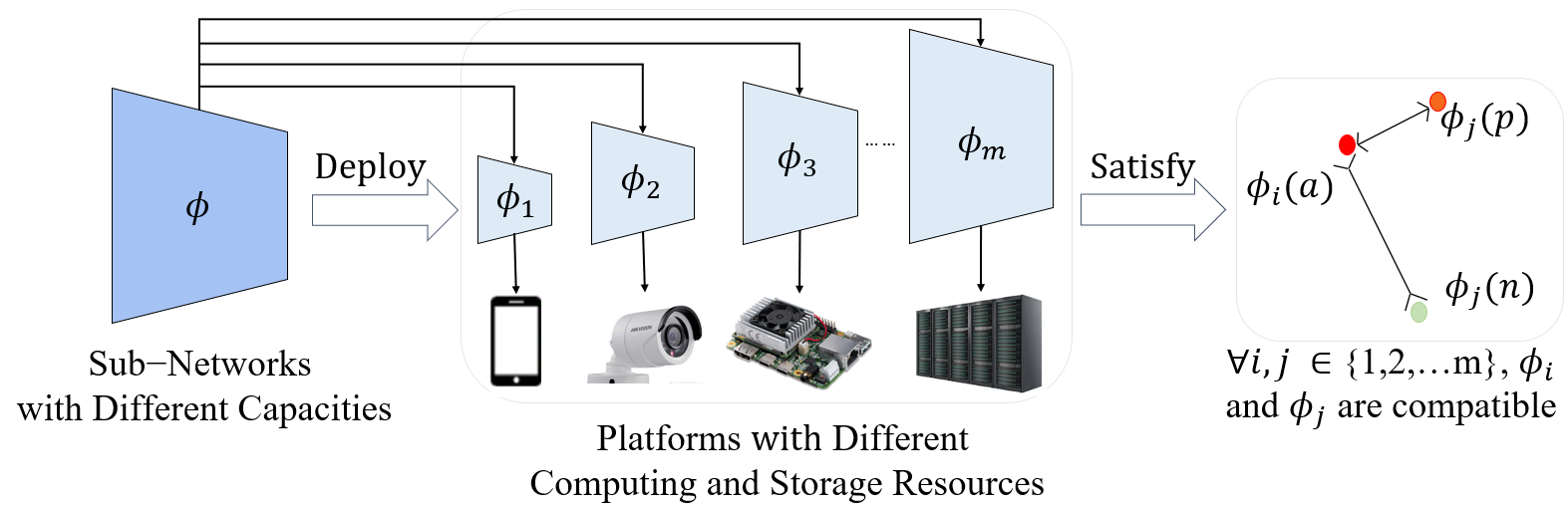}
  \caption{An example of SFSC in multi-platform deployments. The switchable neural network generates a series of sub-networks $\{\phi_1,\phi_2,...,\phi_m\}$ with different capacities for different computing resources. $\forall i,j \in \{1,2,...,m\}$, $|\phi_i(a)-\phi_j(p)|<|\phi_i(a)-\phi_j(n)|$, where $(a,p)$ is a pair of samples with the same class ID, while $(a,n)$ is a pair of samples with different class IDs.   }
  \label{fig:example}
\end{figure*}

In this work, we propose a \textbf{S}witchable representation learning \textbf{F}ramework with \textbf{S}elf-\textbf{C}ompatibility (SFSC) to achieve many-to-many compatibilities. As shown in Figure \ref{fig:example}, SFSC can deploy different sub-models to suit the diverse computing and storage resource limitation of different platforms. The compatibility between any two sub-models can be achieved, termed as ``self-compatibility''. However, there are gradients conflicts between sub-models as they are co-optimized during the training process. 
We summarize such conflicts into the gradient magnitude and gradient direction. Specifically, a gradient of a large magnitude will dominate the optimization process, which may lead to the overfitting of its corresponding sub-model and impair the improvements of other sub-models.  %
To solve this problem, we estimate the uncertainty of different sub-models to weight the gradient, denoting the optimization priorities. Besides, as the sub-models may produce gradients in various directions, there is mutual interference between them. Thus the improvement in different sub-models may be overestimated or underestimated. To tackle such conflict, the gradients are projected onto planes that are orthogonal to each other.

The contributions of this paper are summarized as follows:
\begin{itemize}
    \item We propose a switchable representation learning framework with self-compatibility (SFSC). SFSC generates a series of feature-compatible sub-models with different capacities for visual search systems, which can be deployed on different platforms flexibly.

    \item We resolve the conflicts between sub-models from the aspect of gradient magnitude and gradient direction. A compatible loss based on uncertainty estimation is proposed to guide optimization priorities and alleviate the imbalance of gradient magnitude between sub-models. An aggregation method based on the gradient projection is proposed to avoid mutual interference and find a generic optimal direction for all sub-models. 
    
    \item SFSC achieves state-of-the-art performance on the evaluated benchmark datasets. Compared to deploying a unified model, adopting SFSC to obtain different sub-models can achieve 6\% to 8\% performance improvements on three different datasets. 
    
\end{itemize}

\section{Related Work}

\subsection{One-to-one Compatible Learning Paradigm}

Given a fixed old model, the one-to-one compatible learning methods aim at aligning the learnable models with the fixed old one in metric space, termed as ``backward compatible learning''. 
Existing methods focus on the optimization difficulties caused by the old model. 
Yan \emph{et al.} \cite{yan2021positive} propose Focal Distillation to reduce the ``negative flips'' which means a test sample is correctly classified by the old model but incorrectly classified by the new one. 
Wu \emph{et al.} \cite{wu2021neighborhood} start from neighborhood structures and classification entropy to eliminate the effect of outliers in old features. Shen \emph{et al.} \cite{shen2020non} propose a pseudo classifier and enhance it with a random walk algorithm to get rid of the old classifier. Different from these one-to-one compatible learning methods, this paper studies a new compatible learning paradigm that aims to learn many-to-many compatibility among multiple learnable models.

\subsection{Dynamic Neural Networks}

Dynamic neural networks are able to adapt their structures or parameters to the input during inference \cite{han2021dynamic}. In recent years, research in this field has mainly focused on dynamic depth \cite{graves2016adaptive,huang2017multi,teerapittayanon2016branchynet,bolukbasi2017adaptive,wang2018skipnet,veit2018convolutional}, dynamic width \cite{bengio2013estimating,cho2014exponentially,bengio2015conditional,jacobs1991adaptive,huang2018condensenet,liu2017learning}, dynamic routing \cite{hehn2020end,hazimeh2020tree,yan2015hd,ioannou2016decision}, dynamic parameters \cite{dai2017guodong,zhu2019deformable,gao2019deformable,harley2017segmentation,su2019pixel}, etc. In addition to hand-crafted methods, Neural Architecture Search (NAS) method is also adopted to train and optimize multiple sub-nets with various capacities in one-time training \cite{sahni2021compofa, cai2018proxylessnas, tan2019mnasnet, cai2019once}. 
Dynamic neural networks can allocate computations on demand for different platforms by selectively activating model components. They have been used in classification \cite{wang2020glance,mnih2014recurrent,rosenfeld2016visual}, segmentation \cite{li2017not,roy2018concurrent,kong2019pixel}, and detection tasks \cite{dai2017guodong,zhu2019deformable,verelst2020dynamic}.
In general, current dynamic neural networks output with definite semantic information (\emph{e.g.} class id, detection box) which are naturally interoperable. However, for retrieval tasks, the networks output with features, which requires feature compatibility.


\begin{figure*}[htb]
  \centering
   \includegraphics[height=3.6cm]{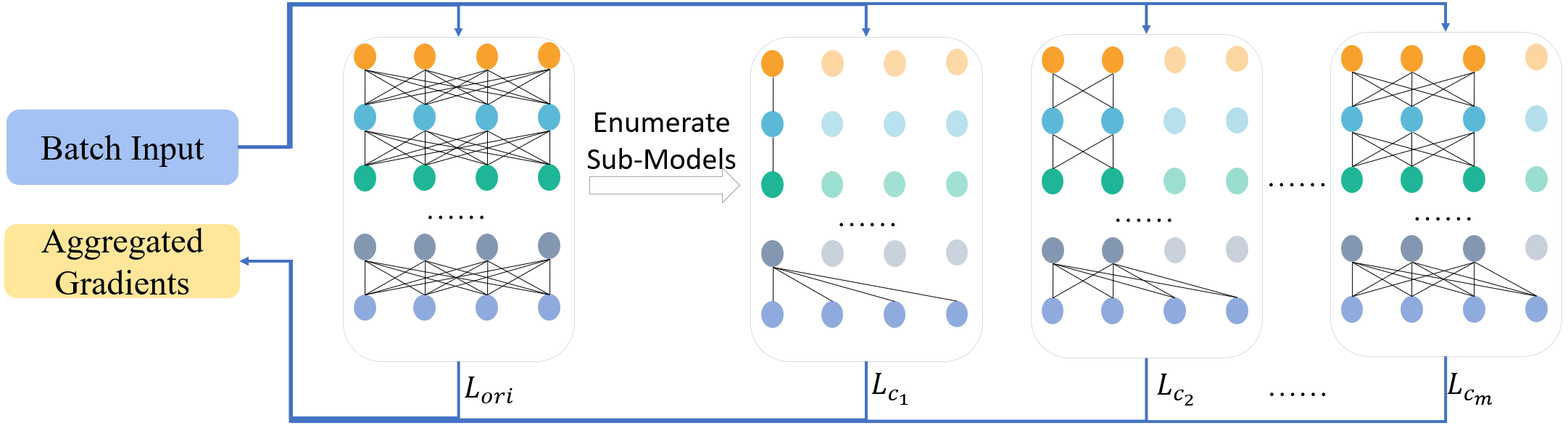}
  \caption{Overview of SFSC. We first transform the full modelcalculate the compatible learning loss $\{ L_{c_1}, L_{c_2},...,L_{c_m}\}$ on different sub-models with various capacitie. $\phi$ into the sub-models. Then we calculate the compatible learning loss $\{ L_{c_1}, L_{c_2},...,L_{c_m}\}$ on different sub-models with various capacities. Finally, we optimize the sub-models by aggregating losses. 
} 
  \label{fig:method}
\end{figure*}

\subsection{Gradient-based Multi-task Optimization}
From the gradient perspective, multi-task optimization is to aggregate gradients corresponding to different tasks and find a descent direction that decreases all objectives. Existing methods treat it as a nonlinear programming problem and take the multi-objective Karush-Kuhn-Tucker \cite{kuhn2014nonlinear} (KKT) conditions to perform gradient projection and rescaling under different assumptions on tasks, \emph{e.g.} multiple tasks which are equally important \cite{sener2018multi,chen2018gradnorm,yu2020gradient}, main tasks with some auxiliary tasks \cite{wang2021gradient,du2018adapting}, a series of the tasks which are learned sequentially \cite{lopez2017gradient,farajtabar2020orthogonal,lin2022trgp}. This paper treats optimizations of multiple sub-models as multiple tasks. As the sub-models vary greatly in their capacity,  the optimization of different models converges at different speeds. The challenge is to dynamically adjust the priorities of different sub-models to avoid the overfitting of someones as well as underfitting of others.

\subsection{Evidential Deep Learning}

Evidential deep learning \cite{EDL} is used to model the uncertainty of deterministic neural networks. Compared with the Bayesian neural networks \cite{mackay1995bayesian,kononenko1989bayesian}, evidential deep learning methods have the advantage of low computational complexity, as there is no need for complex posterior probability estimates and multiple Mento Carlo samplings. Based on the DST \cite{gordon1984dempster} (Dempster-Shafer Theory of Evidence), evidential deep learning treats the output of the neural networks as subjective opinions and formalizes them as Dirichlet distribution. It has been used in classification tasks to overcome the overconfidence problem of softmax cross entropy \cite{BMloss},  action recognition task to identify unknown samples \cite{DEAR}, and multi-view learning task \cite{han2021trusted} to dynamically integrate different views.
This paper treats the output features of different sub-models as evidence to access uncertainty and relates the priority of different sub-models with the obtained uncertainty.

\section{Methods}

In this section, we will describe the proposed SFSC method. We first introduce the construction of switchable neural networks. Then, we analyze the difficulties for optimizations and propose our compatible loss as well as the gradient aggregation method.

\subsection{ Overview}

SFSC involves the training process of multiple sub-models whose features are compatible. Sub-models with different capacities are transformed from the full model $\phi$ according to the crop ratio list $W$. We termed the crop ratio list of each sub-model as $W=\{\gamma_1,\gamma_2,...,\gamma_m\}$, where $\gamma_j$ denotes the ratio of the width of the sub-model $\phi_j$ to the full model $\phi$. 

As shown in Figure \ref{fig:method}, for each batch input, SFSC first calculates the original loss $L_{ori}$. The compatible loss $\{L_{c_1},L_{c_2},...L_{c_m}\}$ is calculated on different sub-models. By aggregating losses $\{L_{ori}, L_{c_1}, L_{c_2},...,L_{c_m}\}$, SFSC computes the general gradient and updates the full model $\phi$ by gradient descent methods.

\subsection{Switchable Neural Network for Sub-models Construction} 
We first convert the traditional convolutional neural network into its corresponding switchable neural network.
As the platforms generally perform model inference layer by layer, the resource limitation is more related to the widths of layers than the number of layers. Here we perform channel pruning on $\phi$ with crop ratio $\gamma_j$ to obtain sub-model $\phi_j$. Specifically, the classic convolutional neural network is mainly composed of some basic modules, such as convolution, fully connected layer, and batch normalization. We modify these basic modules as follows:

1) For convolution (Conv) and fully connected (FC) layers,  if their output is not the final feature for retrieval, we perform pruning in both input and output channels. Otherwise, pruning is only performed in input channels.

2) For batch normalization (BN), we assign an independent BN to the sub-model $\phi_j$. 
As the intermediate features extracted by different sub-models vary greatly in distributions \cite{yu2018slimmable}, channel pruning can not be performed on BN. 
Fortunately, the parameters of BN only occupy a small part of all parameters of the entire network (shown in Table \ref{BNT}). Such an operation only results in a slight increase in storage resources in training progress but does not incur additional deployment costs.

\begin{table}
    \renewcommand{\arraystretch}{1.0}
  \setlength{\tabcolsep}{0.6mm}
    \fontsize{7.5}{8}\selectfont

\label{wip}
  \renewcommand{\arraystretch}{0.6}
\centering
    \fontsize{7}{7.5}\selectfont
\begin{tabular}{l|cccc} \hline
    
    NetWorks & ResNet-18 \cite{he2016identity} & ResNet-50 \cite{he2016identity} & MobileNet V2 \cite{sandler2018mobilenetv2} & ShuffleNet \cite{zhang2018shufflenet}   \\ \hline 

\multirow{2}{*}{Conv and FC}   &   11,679,912& 25,503,912  & 3,470,870  & 5,402,826     \\

&   99.918\%& 99.792\% & 99.027\% & 99.102\% \\\hline

\multirow{2}{*}{BN}   & 9,600& 53,120 & 34,112& 48,960  \\

& 0.0820\%& 0.208\%  & 0.973\%& 0.898\% \\\hline

\hline
\end{tabular}  
\vspace{-0.2cm}
\caption{Number and proportion of different modules' parameters in different network structures}\label{BNT} 

\end{table}

Through such design, we obtain the sub-model $\phi_j$ with the following characteristics: 1) The features extracted by different sub-models have the same dimensions. 2) The ratio of layer widths in the sub-model $\phi_j$ to that in the full model $\phi$ is around $\gamma_j$. 3) The sub-model $\phi_j$ reuses the parameters from the full model $\phi$ except for BN. 

\subsection{Aggregation Method based on the Gradients Projection}

\begin{figure}[!h]
  \centering
   \includegraphics[height=2cm]{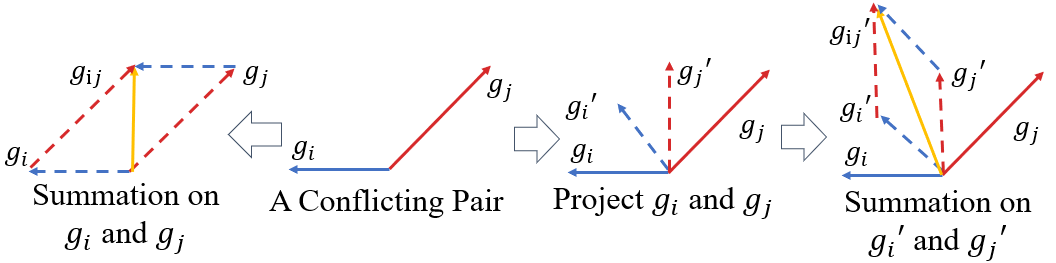}
  \caption{An example of gradient conflict between pair $(g_i,g_j)$. Simply taking the summation aggregation results in the elimination of the improvement of the $i$-th sub-model. Our aggregation module ensures the generic direction to improve both the $i$-th and $j$-th sub-model. } \vspace{-0.5cm}

  \label{fig:vis4}
\end{figure}

 As the gradients corresponding to different sub-models may have different directions, simply aggregating the gradients by summation may cause mutual interference, which means the improvements of sub-models are overestimated or underestimated, as shown in Figure \ref{fig:vis4}.

Inspired by multi-task learning \cite{sener2018multi,bai2020multi,yu2020gradient}, we tackle such a problem by projecting conflicting gradients and keeping the components perpendicular to each other. 
Given a pair of gradients $(g_i,g_j)$ corresponding to sub-model $\phi_i$ and $\phi_j$, if the cosine of their direction $\cos(g_i,g_j) < 0$, they are conflict with each other. For a pair of conflicting gradients $(g_i,g_j)$, we project them to the orthogonal plane of each other, respectively, as shown in Equation \ref{proj}. 

\vspace{-0.6cm}
\begin{align}  
g_j' = g_i -\frac{g_j *g_i} {||g_i||^2}*g_i.
\label{proj}
\end{align}
Such projection results in a direction that improves both the $i$-th and $j$-th sub-model simultaneously.

\subsection{Compatible Loss based on the Uncertainty Estimation}

\begin{figure}[!t]
  \centering
   \includegraphics[height=2cm]{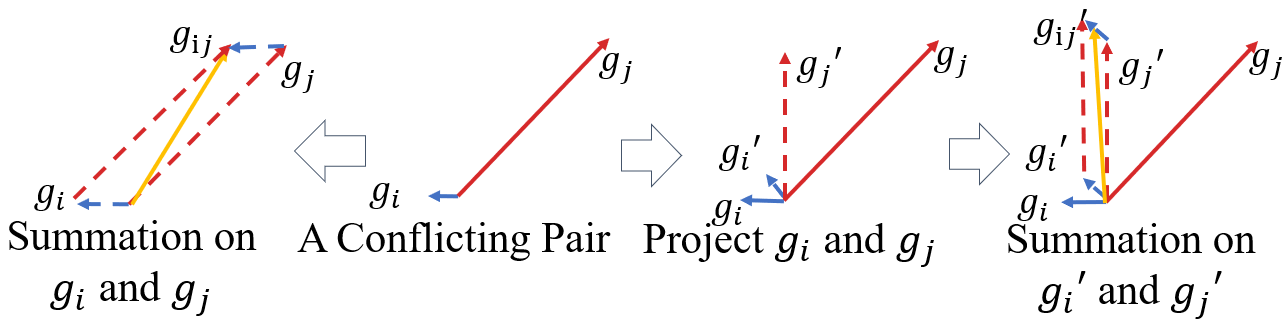}
  \caption{An example of gradient conflict between pair $(g_i,g_j)$ when there is an imbalance of gradient magnitude. The $i$-th sub-model contributes little to the optimization even using gradient projection. } \vspace{-0.5cm}

  \label{fig:vis400}
\end{figure}

The aggregation method aims to optimize all sub-models coordinately. However, it may not work as expected when facing the imbalance of gradient magnitude between sub-models, as shown in Figure \ref{fig:vis400}. Therefore, it is critical to design a compatible loss. 
 
As sub-models are co-optimized, constraining them with existing compatible loss  \cite{shen2020towards,budnik2021asymmetric} faces conflicts in the magnitude of gradients. 
As the sub-models vary in capacities, their optimization process converges at different speeds. The small-capacity sub-models will converge faster. Therefore, the small-capacity sub-model may have reached its performance ceiling while still producing losses of large magnitude, resulting in the overfitting of such sub-model and weakening of other sub-models.  
Intuitively, a sub-model with low uncertainty tends to reach its performance ceiling. Therefore, it is expected to adaptively adjust the optimization priorities of each sub-models through their uncertainty dynamically during the training process.

Here we propose a compatible loss based on the uncertainty estimation. Inspired by Dempster-Shafer Theory of evidence \cite{gordon1984dempster} (DST), we measure the uncertainty through belief, where high uncertainty corresponds to the model with small capacities. Typically, the belief is measured by a set of mass $B=\{b_1,b_2,...b_C\}$ and an overall uncertainty mass $u$, where $b_c$ denotes the belief degree belonging to the class $c$. $B$ and $u$ are related to a Dirichlet distribution, which satisfy Equation \ref{bfmask}. 

\vspace{-0.6cm}
\begin{align}  
u + \sum_{c=1}^{C}b_c =1 , u \ge 0  , b_c \ge 0.
\label{bfmask}
\end{align}
\vspace{-0.4cm}

The belief is estimated by the features extracted by sub-models. Given sample $i$, its features extracted by the $j$-th sub-model termed $\phi_j(i)$. The evidence $E^j_i=\{e_{i1}^j,e_{i2}^j,...e_{iC}^j\}$ is transformed by $\phi_j(i)$ through linear projection and activation function (\textit{i.e., $\mathrm{exp}(\cdot)$}), where the linear projection layers share parameters with the classifier of the $1.0\times$ model $\phi$. The belief of sample $i$ on the $j$-th sub-model is computed by Equation \ref{dilikele}, where $\alpha_{i}^j=\{\alpha_{i1}^j,\alpha_{i2}^j...\alpha_{iC}^j\}$ is the Dirichlet intensity on all classes. 

\vspace{-0.6cm}
\begin{align}  
\begin{split}
&\alpha_{ic}^j = e_{ic}^j+1,\\
S_i^j=\sum_{c=1}^C\alpha_{ic}^j  &, 
b_{ic}^j = \frac{\alpha_{ic}^j-1}{S_i^j}, 
u_i^j = \frac{C}{S_i^j}.
\label{dilikele}
\end{split}
\end{align}
\vspace{-0.4cm}

With $\alpha_{i}^j$, the Dirichlet distribution $\mathrm{D}(p_i|\alpha_i^j)$ is obtained, where $p_i$ is the class assignment probabilities on a simplex. Inspired by the Cross-Entropy loss, we integrate the probability $p_i$ on $\mathrm{D}(p_i|\alpha_i^j)$, as shown in Equation \ref{cce}.

\vspace{-0.6cm}
\begin{align}  
\begin{split}
\mathcal{L}(\alpha_i^j) &= \int [\sum_{c=1}^{C} - y_{ic}^j \log(p_{ic}^j)  ]\frac{1}{\beta(\alpha_i^j)} 	\Pi_{c=1}^C {p_{ic}^j}^{(\alpha_{ic}-1)} {\rm d} p_i^j  \\
&=\sum_{c=1}^C y_{ic}^j(\psi(S_i^j)-\psi(\alpha_{ic}^j  )),
\label{cce}
\end{split}
\end{align}
where $\psi(\cdot)$ is the \textit{digamma} function. Since Equation \ref{cce} may converge to a trivial solution $\mathrm{D}(p_i|1)$ during optimization, we penalize the distance between $\mathrm{D}(p_i|1)$ and $\mathrm{D}(p_i|\alpha_i^j)$ as the regularization. The complete compatible loss $L_{c_j}$ is shown in Equation \ref{s:final}. $L_{c_j}$ adaptively adjusts the magnitude of gradients through their corresponding uncertainty. 

\begin{align}  
\begin{split}
KL\left [  \mathrm{D}(p_i|\alpha_i^j)  ||\mathrm{D}(p_i|\mathbf{1} )\right ] =  \mathrm{log} \left ( \frac{ \sum_{c=1}^C \Gamma(\alpha_{ic}^j)}{\Gamma(C) {\textstyle \prod_{c=1}^{C}(\alpha_{ic}^j )} }  \right )
\\
 +\sum_{c=1}^{C}(\alpha_{ic}^j-1)\left [ \psi(\alpha_{ic}^j -\psi(\sum_{c=1}^C\alpha_{ic}^j))  \right ],\\
 L_{c_j}(i)=\mathcal{L}(\alpha_i^j)+\lambda*KL\left [  \mathrm{D}(p_i|\alpha_i^j)  ||\mathrm{D}(p_i|\mathbf{1} )\right ].
\label{s:final}
\end{split}
\end{align}
\vspace{-0.4cm}

We further conduct a gradient analysis for the proposed compatible loss. The Cross-Entropy loss is written as:

\vspace{-0.6cm}
\begin{align}  
p_{ic}=\frac{e^{z_{ic}}}{\sum_{k=1}^C e^{z_{ik}} },
L_{ce}=\sum_{c=1}^{C}-y_{ic}log{{(p}_{ic})},
\label{CE}
\end{align}
where $z_{i}$ is the classification vector of sample $i$ outputed by the model. 
The corresponding gradient of Equation \ref{CE} can be computed as:
\begin{align}  
 \frac{\partial L_{ce} }{\partial z_{ic} }=\frac{e^{z_{ic}}}{\sum_{k=1}^C e^{z_{ik}} }-y_{ic}.
\label{CEG}
\end{align}
\vspace{-0.2cm}

The corresponding gradient of the proposed compatible loss in \ref{s:final} can be computed as:

\vspace{-0.6cm}
\begin{equation}
\label{eq17}
    \begin{split} 
        \frac{\partial L_{c_j}( \alpha_i^j) }{\partial \alpha_{ic}^j } =\psi^{'}(\alpha_{ic}^j)[  \lambda(\alpha_{ic}^j-1)-\mathbf{y} _{ic}] \\
        +\psi^{'}( S_i^j)[-\lambda( S_i^j-C)+1].
\end{split}
\end{equation}
\vspace{-0.2cm}

We take $\frac{1}{\alpha_{ic}^j}$ as the approximation of $\psi^{'}(\alpha_{ic}^j)$ and obtain Equation \ref{eq18}.

\vspace{-0.6cm}
\begin{equation}
\label{eq18}
    \begin{split} 
        \frac{\partial L_{c_j}( \alpha_i^j) }{\partial \alpha_{ic}^j } \approx \frac{1}{\alpha_{ic}^j}[  \lambda(\alpha_{ic}^j-1)-\mathbf{y} _{ic}] \\ 
        +\frac{1}{ S_i^j}[-\lambda( S_i^j-C)+1]
        \\= - (\frac{\lambda+y_{ic}}{\alpha_{ic}^j}+\frac{\lambda C+1}{S_i^j}),
\end{split}
\end{equation}
where the first term corresponds to the Cross-Entropy loss in Equation \ref{CEG}, and the second term corresponds to the uncertainty estimation. As the uncertainty $u_i^j = \frac{C}{S_i^j} \propto\frac{1}{S_i^j}  $, thus higher $u_i^j$ would result in a gradient with larger magnitude through $\frac{\lambda C+1}{S_i^j}$. Therefore, the magnitude of gradients corresponding to different sub-models is 
dynamically adjusted to avoid someone dominating the optimization process.

Such compatible loss can work well with gradient projection, as shown in Algorithm \ref{AAGP}. 
\begin{algorithm}[H]
  \fontsize{9}{9}\selectfont
    \caption{Training Process of SFSC}\label{AAGP} 
    \begin{algorithmic}[1]
        \Require batch input $\mathcal{B}$, the crop list $W=\{\gamma_0,\gamma_1,...,\gamma_m\}$, the full model $\phi$
            \State $g_{ori} \gets \frac{\delta L_{ori}(\phi, \mathcal{B}) }{\delta \phi}$
            \For {$ \gamma_j \in\ W  $}
                \State Switch to the sub-model $\phi_j$
                \State $g_{j} \gets \frac{\delta L_{c_j}(\phi_j, \mathcal{B}) }{\delta \phi}$
            \EndFor
            \State $G, G^p \gets \{g_{ori},g_0,...g_m\}$
            \For {$ g_a \in\ G^p  $}
                \State  Shuffle(G)
                \For {$ g_b \in\ G  $}
                    \State $d = g_a *g_b$
                    \If {$d < 0$ } //have conflicts
                        \State $g_a \gets g_a- \frac{g_b*d}{||g_b||^2}$
                    \EndIf
                \EndFor
            \EndFor
    \State \textbf{return} Update $\phi$ by $\Delta \theta= - \sum_{g_a \in G^p} g_a$
    \end{algorithmic}
\end{algorithm}

In general, SFSC aims to achieve the Pareto optimality \cite{censor1977pareto} among the optimization of sub-models. The proposed compatible loss adaptively adjusts the magnitude of gradients through uncertainty to control the optimization priorities of sub-models dynamically. Besides, the aggregation method performs the gradient projection to find a generic direction to improve all sub-models. 

\begin{table*}[!t]
\vspace{-0.3cm}
  \centering
  \setlength{\tabcolsep}{2.6mm}  
  \renewcommand{\arraystretch}{1.2}
    \fontsize{9}{9}\selectfont

  \label{b:baseline2}
\begin{tabular}{c|c|c|c|c||c|c|c|c}
\hline
\multicolumn{1}{l|}{\diagbox{ $\phi_q$ } {$M(\phi_{q},\phi_{g})$} { $\phi_g$ }} & 

\multicolumn{1}{c|}{ $\phi_{\frac{1}{16}}$   } & \multicolumn{1}{c|}{$\phi_{\frac{1}{4}}$  } & \multicolumn{1}{c|}{$\phi_{\frac{9}{16}}$   } & \multicolumn{1}{c||}{$\phi$ } &\multicolumn{1}{c|}{ $\phi_{\frac{1}{16}}$   } & \multicolumn{1}{c|}{$\phi_{\frac{1}{4}}$  } & \multicolumn{1}{c|}{$\phi_{\frac{9}{16}}$   } & \multicolumn{1}{c}{$\phi$ }   \\\hline 

\multicolumn{1}{l|}{\textbf{}} & \multicolumn{4}{c||}{Unified Model}                                                                                                           & \multicolumn{4}{c}{Ours}       \\ \hline
$\phi_{\frac{1}{16}}$          & 55.25 &   -    &   -    &      - & 58.19   & 62.88  & 63.48 & 69.43 \\
$\phi_{\frac{1}{4}}$                   &    -   & 67.48 &     -  &   -   & 61.24   & 70.74  & 71.37 & 76.37 \\
$\phi_{\frac{9}{16}}$               &   -    &      - & 71.25 &  -     & 62.43   & 71.25  & 72.06 & 77.26 \\
$\phi$                    &  -     &   -    &   -    & 80.91 & 68.03   & 74.76  & 75.67 & 81.43 \\ \hline \hline

\multicolumn{1}{l}{} & \multicolumn{4}{c||}{BCT-S}     & \multicolumn{4}{c}{Asymmetric-S} \\ \hline
$\phi_{\frac{1}{16}}$                   & 54.40  & 58.10  & 60.22 & 61.29 & 48.79   & 51.96  & 54.09 & 56.31 \\
$\phi_{\frac{1}{4}}$                   & 55.71 & 66.38 & 68.29 & 69.71 & 52.24   & 62.45  & 64.98 & 67.23 \\
$\phi_{\frac{9}{16}}$                    & 56.50  & 68.30  & 68.42 & 71.99 & 54.44   & 64.74  & 67.97 & 71.28 \\
$\phi$                           & 61.79 & 69.73 & 71.93 & 73.61 & 56.54   & 67.30   & 71.91 & 78.38\\ \hline\hline

\multicolumn{1}{l}{} & \multicolumn{4}{c||}{BCT}       & \multicolumn{4}{c}{Asymmetric}   \\ \hline 
$\phi_{\frac{1}{16}}$                 & 55.55 & 56.64 & 59.40  & 60.35 & 55.83   & 50.85  & 53.71 & 55.39 \\
$\phi_{\frac{1}{4}}$                    & 55.83 & 65.74 & 67.10  & 68.54 & 52.66   & 67.04  & 62.08 & 66.59 \\
$\phi_{\frac{9}{16}}$                      & 55.66 & 68.02 & 67.49 & 70.93 & 54.48   & 62.33  & 66.33 & 70.14 \\
$\phi$                            & 58.69 & 69.48 & 70.04 &  -     & 56.08   & 66.66  & 69.71 &   -    \\\hline \hline 

\end{tabular}  
\vspace{-0.2cm}
  \caption{Baseline performance comparison on Market1501 (mAP). ``Unified Model'' are models without any compatible regularization. }
\end{table*}

\begin{table*}[!t]
  \centering
  
  \setlength{\tabcolsep}{3.3mm}
    \renewcommand{\arraystretch}{1.2}
    \fontsize{9}{9}\selectfont
\begin{tabular}{l|c|c|c|c|c|c|c|c} \hline
    ~ & \multicolumn{2}{c|}{\textbf{$M(\phi_{\frac{1}{16}},\phi)$ }} &	\multicolumn{2}{c|}{\textbf{$M(\phi_{\frac{1}{4}},\phi)$  }}&	\multicolumn{2}{c|}{\textbf{$M(\phi_{\frac{9}{16}},\phi)$  }}&	\multicolumn{2}{c}{$M(\phi,\phi)$ }  \cr \cline{2-9}
      ~ &R1 &mAP &R1 &mAP &R1 &mAP &R1 &mAP \\ \hline \hline

     \multicolumn{9}{c}{MSMT17} \\ \hline 
Unified Model & 28.82                & 11.69                & 48.65                & 22.86                & 57.56                & 30.06                & 70.22                & 43.06                \\ \hline
BCT-S             & 46.49                & 20.11                & 56.22                & 28.13                & 58.11                & 30.14                & 58.76                & 30.99                \\
Asymmetric-S            & 32.64                & 13.34                & 49.62                & 24.04                & 57.18                & 30.04                & 66.08                & 40.39                \\
Ours              & \textbf{56.22}       & \textbf{28.16}       & \textbf{63.29}       & \textbf{35.32}       & \textbf{65.13}       & \textbf{37.74}       & \textbf{70.38}       & \textbf{43.89}       \\\hline \hline

     \multicolumn{9}{c}{VERI-776} \\ \hline 
Unified Model&	81.07&	44.79&	85.30 &	53.63& 88.69 &55.42&92.20 &66.50\\ \hline
BCT-S&	\textbf{86.25}&	49.38&	87.08&	57.03& 90.42 &58.20 &86.73	&	58.46    \\
Asymmetric-S& 68.75&	42.16&	84.29&	56.55 &	88.04&	61.37&	89.82 &	65.35 \\
Ours&	79.76&	\textbf{55.04}&	\textbf{89.94}&	\textbf{62.28} & \textbf{90.48}&	\textbf{62.72}&	\textbf{92.32}&	\textbf{66.55}\\\hline \hline
\end{tabular}       \vspace{-0.3cm} 
  \caption{Performance comparison on MSMT17 and VeRi-776 datasets.}  \label{b:baseline}

\end{table*}

\begin{table*}[!t]
  \centering
  \setlength{\tabcolsep}{3.3mm}
    \renewcommand{\arraystretch}{1.2}
    \fontsize{9}{9}\selectfont
\begin{tabular}{l|c|c|c|c|c|c|c|c} \hline
    ~ & \multicolumn{2}{c|}{\textbf{$M(\phi_{\frac{1}{16}},\phi)$ }} &	\multicolumn{2}{c|}{\textbf{$M(\phi_{\frac{1}{4}},\phi)$  }}&	\multicolumn{2}{c|}{\textbf{$M(\phi_{\frac{9}{16}},\phi)$  }}&	\multicolumn{2}{c}{$M(\phi,\phi)$ }  \cr \cline{2-9}
      ~ &R1 &mAP &R1 &mAP &R1 &mAP &R1 &mAP \\ \hline \hline

\multicolumn{9}{c}{Circle Loss}                                                                                                                      \\  \hline

Unified Model & 70.25          & 50.79          & 72.66          & 60.43          & 78.21          & 64.66          & 87.38          & 74.06          \\ \hline
BCT-S          & 63.18          & 42.45          & 72.74          & 52.04          & 77.38          & 59.49          & 87.53          & 74.58          \\
Asymmetric-S       & 69.66          & 48.21          & 77.85          & 57.02          & 80.52          & 61.81          & 86.90          & 73.08          \\
Ours         & \textbf{73.99} & \textbf{54.61} & \textbf{80.58} & \textbf{63.46} & \textbf{82.39} & \textbf{67.02} & \textbf{88.95} & \textbf{77.05} \\  \hline\hline 
\multicolumn{9}{c}{Softmax   + Triplet}                                                                                                              \\\hline 
Unified Model         & 77.55          & 55.82          & 83.97          & 65.30           & 87.89          & 70.69          & 90.94          & 78.30          \\\hline
BCT-S          & 81.62          & 59.58          & 87.86          & 69.05          & 88.75          & 71.38          & 90.11          & 74.71          \\
Asymmetric-S         & 75.56          & 52.60          & 85.36          & 64.21          & 86.91          & 68.83          & 90.45          & 77.34          \\
Ours         & \textbf{87.36} & \textbf{66.21} & \textbf{90.94} & \textbf{76.64} & \textbf{91.21} & \textbf{77.93} & \textbf{92.64} & \textbf{81.46} \\\hline \hline 

\end{tabular}
     \vspace{-0.3cm} 
  \caption{Performance comparison between different loss functions
on Market1501.}  \label{b:losschg}

\end{table*}

\begin{table*}
  \centering
  \renewcommand{\arraystretch}{1.2}
  \setlength{\tabcolsep}{3.3mm}
    \fontsize{9}{9}\selectfont

\begin{tabular}{l|c|c|c|c|c|c|c|c} \hline
    ~ & \multicolumn{2}{c|}{\textbf{$M(\phi_{\frac{1}{16}},\phi)$ }} &	\multicolumn{2}{c|}{\textbf{$M(\phi_{\frac{1}{4}},\phi)$  }}&	\multicolumn{2}{c|}{\textbf{$M(\phi_{\frac{9}{16}},\phi)$  }}&	\multicolumn{2}{c}{$M(\phi,\phi)$ }  \cr \cline{2-9}
      ~ &R1 &mAP &R1 &mAP &R1 &mAP &R1 &mAP \\ \hline \hline

\multicolumn{9}{c}{ResNet-50}                                                                                                                      \\  \hline
Unified Model & 85.27          & 65.44          & 88.84          & 72.07          & 90.8           & 76.40          & 92.22          & 82.13          \\  \hline
BCT-S           & 89.31          & 72.09          & 91.00             & 77.22          & 91.36          & 78.27          & 91.86          & 79.40           \\
Asymmetric-S          & 85.04 & 67.39 & 90.26 & 76.28 & 91.60  & 78.53 & 91.95 & 82.79 \\
Ours            & \textbf{90.41} & \textbf{76.40}  & \textbf{91.03} & \textbf{78.20}  & \textbf{91.66} & \textbf{78.76} & \textbf{93.44} & \textbf{83.66} \\ \hline \hline

\multicolumn{9}{c}{MobileNet V2}                                                                                                                        \\  \hline
Unified Model & 69.00             & 41.14          & 89.58          & 72.85          & 91.54          & 77.99          & 92.28          & 79.50           \\  \hline
BCT-S           & 80.46          & 55.17          & 88.48          & 70.68          & 89.88          & 73.72          & 89.96          & 73.87          \\
Asymmetric-S          & 67.28          & 43.64          & 86.97          & 68.95          & 89.31          & 73.82          & 90.74          & 78.45          \\
Ours            & \textbf{83.11} & \textbf{63.82} & \textbf{90.53} & \textbf{76.52} & \textbf{91.18} & \textbf{78.30}  & \textbf{92.61} & \textbf{81.34} \\ \hline \hline
\multicolumn{9}{c}{ShuffleNet}                                                                                                                         \\ \hline
Unified Model & 82.54          & 60.12          & 89.85                  & 64.45          & 91.95    & 72.96          & 84.38        & 79.13          \\ \hline
BCT-S           & 85.51          & 65.76          & 88.42                & 66.85          & 89.73    & 70.59          & 86.67          & 74.41          \\
Asymmetric-S          & 82.30          & 62.50           & 84.41                   & 57.43          & 88.42   & 66.43          & 78.92        & 74.92          \\
Ours            & \textbf{89.25} & \textbf{73.09}  & \textbf{89.93} & \textbf{74.86}& \textbf{89.93} & \textbf{75.14} & \textbf{92.81} & \textbf{81.97} \\ \hline \hline
\end{tabular}
     \vspace{-0.3cm} 
  \caption{Performance comparison between different model architectures
on Market1501.}  \label{b:modelchg}
\end{table*}

\section{Experiments}

\subsection{Experimental Configuration}

\textbf{Datasets}. We evaluate the proposed SFSC on person ReID datasets, \textit{i.e.}, Market-1501~\cite{zheng2015scalable}, MSMT17~\cite{wei2018person}, and one vehicle ReID dataset VeRi-776~\cite{wang2017orientation}. Market1501 consists of 32,668 annotated images of 1,501 identities shot from 6 cameras. MSMT17 consists of 126,441 bounding boxes of 4,101 identities taken by 15 cameras. VeRi-776 contains over 50,000 images of 776 vehicles captured by 20 cameras.

\textbf{Comparison methods}. For the one-to-one compatible learning paradigm, BCT \cite{shen2020towards} and Asymmetric \cite{budnik2021asymmetric} are adopted, as they are representatives of compatible learning through classification losses and embedding losses.  BCT and Asymmetric can only obtain sub-models in a one-to-one compatible way. Specifically, given the fixed largest capacity model which is trained without compatible regularization, we train a series of small-capacity models to be compatible with it, respectively.
Besides, we also incorporate BCT/Asymmetric loss in our proposed self-compatible learning paradigm for a fair comparison. We take the compatible loss in BCT/Asymmetric to constrain each sub-model compatible with the full model, and use summation to aggregate losses from sub-models, termed as ``BCT-S'' and ``Asymmetric-S''. 

\textbf{Evaluation metrics}. Mean average precision (mAP) and top-1 accuracy (R1) are adopted to evaluate the performances of retrieval. Define $M(\phi_a, \phi_b)$ as the performances of retrieval where the query features are extracted by model $\phi_a$ and gallery features are extracted by model $\phi_b$. We set $\phi_a$ and $\phi_b$ as different sub-models to evaluate the performance of compatible learning.

\textbf{Implementation details}. The implementations of all compared methods and SFSC are based on the FastReID~\cite{he2020fastreid}. The default configuration named ``bag of tricks'' \cite{luo2019bag} is adopted. $\lambda$ is set to 0.2 for all experiments.

\subsection{Baseline Comparisons}



\begin{figure*}[!t]
  \centering
   \includegraphics[height=5.0cm]{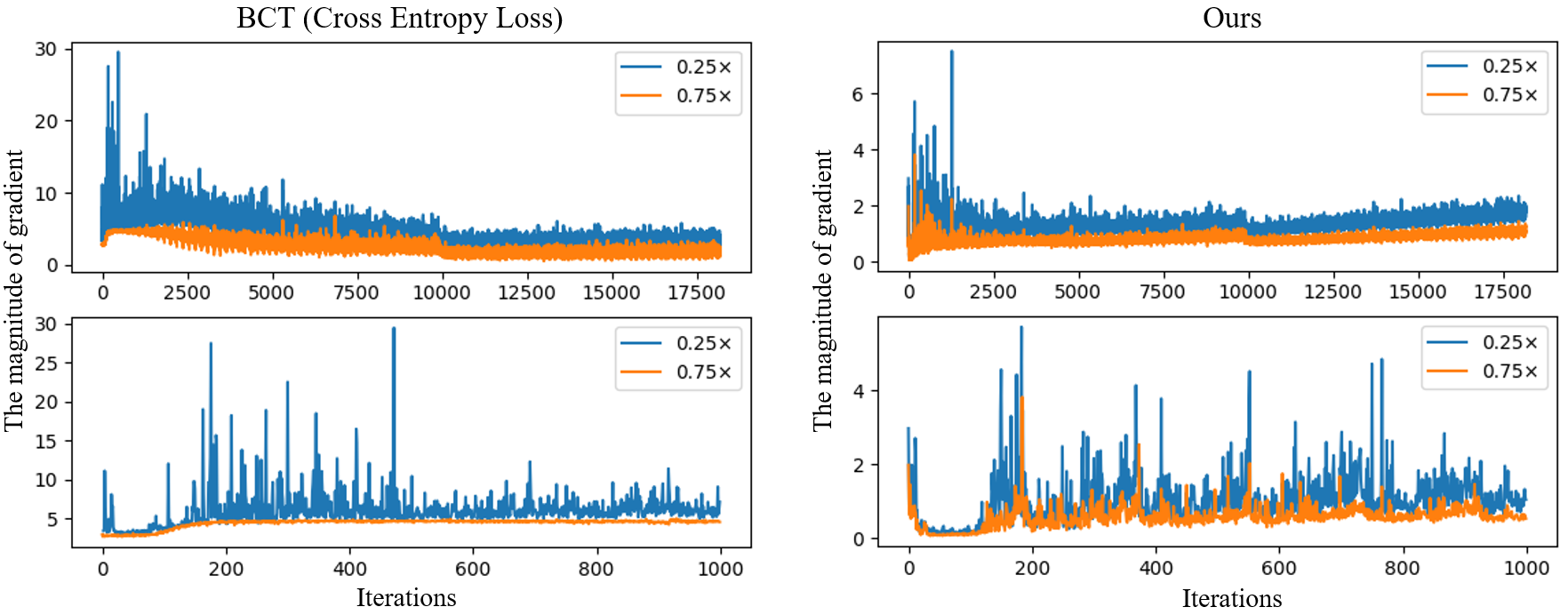}
  \caption{Gradient visualization on Market1501-R18 baseline. As for the original cross-entropy loss (BCT \cite{shen2020towards}), the magnitude of the $0.25\times$ sub-model is much larger than that of the $0.75\times$ sub-model, especially at the early stage of the training process (0$\sim$1000 iterations). Therefore, the $0.25\times$ sub-model dominates the training progress and the improvements of the $0.75\times$ sub-models are eliminated. While our method ensures co-optimization through uncertainty estimation. } \vspace{-0.3cm}
  \label{fig:show} 
\end{figure*}


We conduct baseline experiments on Market1501. We take ``Softmax + Cross Entropy'' as the loss function, ResNet-18 \cite{he2016identity} as the backbone, and $W=\{0.25\times, 0.50\times, 0.75 \times\}$ as the crop ratio list of the sub-models. Notes that for a model with crop ratio $\gamma_i$, the ratio of its parameters to the full model is $\gamma_i^2$. Therefore, the ratio of the number of parameters in the $0.25\times$, $0.50\times$ and $0.75 \times$ sub-model  to that in the full model are $\frac{1}{16}$, $\frac{1}{4}$, $\frac{9}{16}$, respectively.

\textbf{Comparison methods}. As for one-to-one compatible learning methods BCT and Asymmetric, the compatibility does not outperform that achieved by their corresponding self-compatible methods especially when the models are not constrained to be compatible directly. For self-compatible learning methods, BCT-S and Asymmetric-S, the conflicts among sub-models lead to a limited average accuracy.

\textbf{Our Method (SFSC)}. As we resolve the conflicts among different sub-models, our methods can achieve many-to-many compatibility among all the sub-models, which significantly outperforms all comparison methods.

\subsection{Performances under Different Settings}

\textbf{Different datasets}. We evaluate each self-compatible learning method on the other person ReID datasets MSMT17, and the vehicle ReID dataset VeRi-776 under baseline configurations, as shown in Table \ref{b:baseline}. The results on the two datasets with larger scales are consistent with that on the Market1501. Our methods outperform others significantly for the small capacity sub-model (\emph{e.g.} the $0.25\times$  sub-model $\phi_{\frac{1}{4}}$).

\textbf{Different loss functions}. We replace the original loss function $L_{ori}$ from ``Softmax'' to ``Softmax + Triplet'' and ``Circle Loss'' \cite{sun2020circle} to evaluate the effectiveness of each method, while Triplet loss is a kind embedding loss and circle loss is a kind of classification loss. The results in Table \ref{b:losschg} show that the accuracy of BCT-S and Asymmetric-S drops greatly when training with circle loss, as  Our method still stably achieves the best performance.

\textbf{Different model architectures}. We also study whether each method can be stably applied to various model architectures, \textit{i.e.}, ResNet-50 \cite{he2016identity}, MobileNet V2 \cite{sandler2018mobilenetv2} and ShuffleNet \cite{zhang2018shufflenet}. ShuffleNet and MobileNet are both designed especially for edge devices, which meet the application scenarios of self-compatible learning. It is shown in Table \ref{b:modelchg} that BCT-S and Asymmetirc-S do not achieve higher accuracy than deploying a unified model with ShuffleNet and MobileNet (\emph{e.g.} for the $0.75\times$ sub-model $\phi_{\frac{9}{16}}$). And our methods always improve the accuracy in all cases.

\subsection{Ablation Study}

We replace the proposed compatible loss and aggregation module with existing compatible loss BCT \cite{Shen2020} and summation aggregation to test the proposed two methods. 
\begin{table}[!h]
  \centering
    \renewcommand{\arraystretch}{1.0}
  \setlength{\tabcolsep}{1.0mm}
    \fontsize{7.5}{8}\selectfont
\begin{tabular}{l|c|c|c|c} \hline
    ~  &\textbf{$M(\phi_{\frac{1}{16}},\phi)$} &	\textbf{$M(\phi_{\frac{1}{4}},\phi)$}&	\textbf{$M(\phi_{\frac{9}{16}},\phi)$}&	\textbf{$M(\phi,\phi)$} \\\hline
Unified Model &	55.25&	67.48&	71.25&	80.91\\\hline
w/o Ours Loss&	66.41&	74.58&	76.86&	77.87\\
w/o Ours Aggregation
&	68.08&	73.76&	74.84&	75.38\\ 
SFSC &	\textbf{69.43}&	\textbf{76.37}&	\textbf{77.26}&	\textbf{81.43}\\\hline

\end{tabular}  \vspace{-0.3cm}
  \caption{Ablation Study on Market1501.}
  \label{b:5.6}
\end{table}

Table \ref{b:5.6} shows that the proposed compatible loss mainly improves the retrieval accuracy of the small capacity sub-models while the proposed aggregation method can ensure the performance of the full model.

\subsection{Hyperparameter Analysis}

Here we verify the impact of hyperparameter settings on the performance of the proposed method. The key hyperparameter is the crop ratio list for sub-models $W$. Intuitively, if the number of $W$ increases or the minimal crop ratio decreases, the compatibility constraints on model $\phi$ would be more strict, thereby increasing the difficulty of compatible learning. Therefore, we take the baseline configuration and set $W$ with different values.

\begin{table}[!h]
\vspace{-0.1cm}
  \centering
  \renewcommand{\arraystretch}{1.0}
  \setlength{\tabcolsep}{0.2mm}
    \fontsize{7}{7.5}\selectfont
\begin{tabular}{l|c|c|c|c|c} \hline
~&	 \textbf{$M(\phi_{\frac{1}{16}},\phi)$}&	\textbf{$M(\phi_{\frac{81}{400}},\phi)$}&	\textbf{$M(\phi_{\frac{169}{400}},\phi)$}&	\textbf{$M(\phi_{\frac{289}{400}},\phi)$}&	\textbf{$M(\phi,\phi)$} \\\hline
Unified Model&	55.25&	62.54&	69.54&	72.57&	80.27\\\hline
BCT-S&	61.45&	68.63&	73.42&	72.80&	75.39\\
Asymmetric-S&	62.87&	69.45&	75.20&	73.14&	74.60\\
Ours&	\textbf{69.63}&	\textbf{75.81}&	\textbf{77.10}&	\textbf{77.36}&	\textbf{80.59}\\\hline
\end{tabular}          \vspace{-0.3cm} 
  \caption{Performance comparison with $W$$=$$\{0.25\times,0.45\times,0.65\times,0.85\times\}$ on Market1501 (mAP).}
  \label{b:5.7}\vspace{-0.5cm}
\end{table}

\begin{table}[!h]
  \centering
  \renewcommand{\arraystretch}{1.0}
  \setlength{\tabcolsep}{1.6mm}
    \fontsize{7.5}{8}\selectfont
\begin{tabular}{l|c|c|c|c} \hline
    ~ & \textbf{$M(\phi_{\frac{1}{100}},\phi)$} &	\textbf{$M(\phi_{\frac{1}{16}},\phi)$}&	\textbf{$M(\phi_{\frac{9}{16}},\phi)$}&	\textbf{$M(\phi,\phi)$}\\\hline
Unified Model &	40.40&	55.25&	71.25&	80.27 \\\hline
BCT-S&	42.34	&59.75&	71.73&	76.39 \\
Asymmetric-S&	42.58&	58.20&	70.50&	75.42 \\
Ours&	\textbf{45.44} &	\textbf{68.18}&	\textbf{78.00}&	\textbf{81.46} \\\hline

\end{tabular}          \vspace{-0.3cm} 
  \caption{Performance comparison with $W$$=$$\{0.10\times,0.25\times,0.75\times\}$ on Market1501 (mAP).}
  \label{b:5.8}\vspace{-0.4cm}
\end{table}


Table \ref{b:5.7} shows the performances with the number of $W$ increases, while Table \ref{b:5.8} shows the performances with the minimum among $W$ decreases. The result of the comparison methods has fluctuated to varying degrees. SFSC still achieves the best retrieval accuracy.

\subsection{Visualization}

\begin{figure}[!t]
  \centering
   \includegraphics[height=5.0cm]{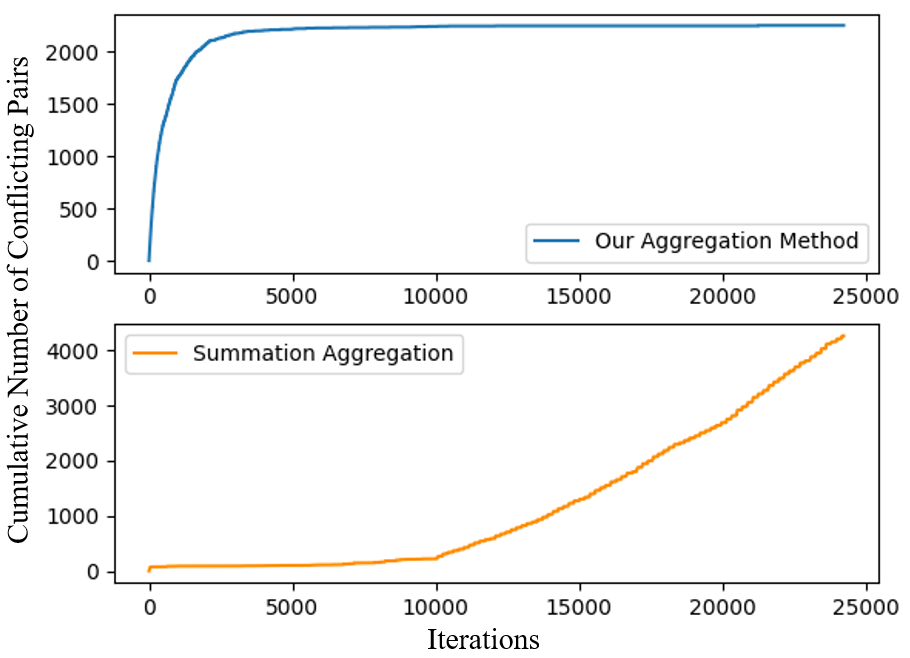}
   \vspace{-0.3cm}
  \caption{The cumulative number of conflict pairs during the training process. 
  }
  \label{fig:conf}\vspace{-0.5cm}
\end{figure}

We visualize the norm of the gradient during the training progress on Market1501-R18 baseline, as shown in Figure \ref{fig:show}. We also visualize the cumulative number of gradient pairs which are conflict with each other during the training process. As shown in Figure \ref{fig:conf}, our method has found a generic direction to improve all sub-models. Therefore, there are almost no conflict pairs after 5000 iterations. As for the summation aggregation, it faces fierce conflicts especially after 10000 iterations, resulting in the improvement of different sub-models overestimated or underestimated.

\section{Conclusion}

This paper proposes a Switchable Representation Learning Framework with Self Compatibility (SFSC) for multi-platform model collaboration. SFSC enables us to obtain models with various capacities to fit different computing and storage resource constraints on diverse platforms. It finds a generic direction to improve all sub-models by uncertainty estimation and gradient projection.
SFSC achieves state-of-the-art performances and shows convincing robustness in different case studies.

\section{Acknowledgments}
This work was supported by the National Natural Science Foundation of China under Grant 62088102, and in part by the PKU-NTU Joint Research Institute (JRI) sponsored by a donation from the Ng Teng Fong Charitable Foundation.

\newpage\clearpage

{\small
\bibliographystyle{ieee_fullname}
\bibliography{egbib}
}

\end{document}